# Volume-Wise Task fMRI Decoding with Deep Learning: Enhancing Temporal Resolution and Cognitive Function Analysis

Yueyang Wu, Sinan Yang, Yanming Wang, Jiajie He, Muhammad Mohsin Pathan, Bensheng Qiu, and Xiaoxiao Wang

*Abstract*—In recent years, the application of deep learning in task functional Magnetic Resonance Imaging (tfMRI) decoding has led to significant advancements. However, most studies remain constrained by assumption of temporal stationarity in neural activity, resulting in predominantly block-wise analysis with limited temporal resolution on the order of tens of seconds. This limitation restricts the ability to decode cognitive functions in detail. To address these limitations, this study proposes a deep neural network designed for volume-wise identification of task states within tfMRI data, thereby overcoming the constraints of conventional methods. Evaluated on Human Connectome Project (HCP) motor and gambling tfMRI datasets, the model achieved impressive mean accuracy rates of 94.0±8.2% and 79.6±7.1%, respectively. These results demonstrate a substantial enhancement in temporal resolution, enabling more detailed exploration of cognitive processes. The study further employs visualization algorithms to investigate dynamic brain mappings during different tasks, marking a significant step forward in deep learning-based frame-level tfMRI decoding. This approach offers new methodologies and tools for examining dynamic changes in brain activities and understanding the underlying cognitive mechanisms.

## I. INTRODUCTION

FUNCTIONAL Magnetic Resonance Imaging (fMRI) is a non-invasive brain imaging technique that measures changes in Blood Oxygen Level Dependent (BOLD) signals over time. By decoding fMRI data, it is possible to gain valuable insights into the cognitive states of the human brain. fMRI possesses high spatial resolution and relatively good temporal resolution [1], [2], [3], [4], [5], enabling precise localization of activities in different brain regions and capturing transient neural activity patterns [6]. Moreover, this technology contributes to developing brain-computer interfaces (BCIs) by decoding text or other outputs from neural signals, bridging cognitive neuroscience with practical applications [7]. Over the past decade, deep learning has revolutionized neuroimaging analysis by enabling high-accuracy models [8], [9], [10] to decode cognitive functions and identify biomarkers. In recent years, there has been growing interest in exploring deep learning techniques to improve the decoding of human brain cognitive functions in the field of neuroimaging, with good generalizability and transferability.

The temporal resolution of fMRI deep learning decoding is fundamentally constrained by the intrinsic characteristics of the hemodynamic response function (HRF). The HRF, inherently sluggish with a full-width-at-half-maximum (FWHM) around 5 seconds [11], blurs the temporal precision of fMRI signals to around 1 second [12]. It is commonly assumed that the hemodynamic response is a linear transformation of the underlying fast neuronal signal, which may be fitted and detected by the general linear model (GLM) analysis [13]. GLM-based decoding methods, such as sparse dictionary learning approach proposed by Ge et al. [14], have been employed to detect dynamic brain activities, and extend GLM by incorporating manual feature selection and have been widely applied in naturalistic image reconstruction tasks [15]. Despite GLM's utility in post-hoc analysis, its assumptions of linearity and stationarity limit its ability to capture dynamic neural activity [16], [17]. Another common approach is the block-design decoding, which classifies a series of volumes into one label [9]. However, the high temporal resolution volume-wise decoding remains underexplored. For instance, Wang et al. [18] employed a Deep Sparse Recurrent Neural Network (DSRNN) for brain state classification. But this approach flattens spatial data, requires strict preprocessing, and relies on manual subnetwork tuning, which collectively diminish its interpretability and practical applicability.

In this work, we propose an end-to-end, high-precision, interpretable, and cross-subject deep neural network for frame-level decoding. Experimental results demonstrate that the proposed model has achieved commendable

This study was supported by National Science and Technology Innovation 2030 Major Program 2022ZD0204801. Funding supports from the National Key R&D Program of China (grant 2022YFB4702700, G.-Z.Y.)

The data employed in this study were obtained from the publicly available Human Connectome Project (HCP) database, which received ethical approval from the Washington University Institutional Review Board (IRB#201204036).

Yueyang Wu, Sinan Yang, Yanming Wang, Muhammad Mohsin Pathan, Bensheng Qiu, and Xiaoxiao Wang are with the Medical Imaging Center, Department of Electronic Engineering and Information Science, University of Science and Technology of China, Anhui, Hefei 230031, China (e-mail: wuyueyang@mail.ustc.edu.cn; ysnysn@mail.ustc.edu.cn; ming1258@ustc.edu.cn; mohsinpathan@mail.ustc.edu.cn; bqiu@ustc.edu.cn; wang506@ustc.edu.cn).

Jiajie He is with the Department of Computer Science and Electrical Engineering，University of Maryland Baltimore County, Maryland, Baltimore 21250, USA (e-mail: jiajieh1@umbc.edu).

performance on both block-design based task functional Magnetic Resonance Imaging (tfMRI) and event-related design based tfMRI. We also conducted a visualization analysis of the model's results.

## II. METHOD

### A. Data Description

The dataset employed in this study is the tfMRI data from Human Connectome Project (HCP) S1200 release. This dataset includes tfMRI data from 1034 participants across 7 cognitive-related tasks, encompassing both block-design and event-related design paradigms. The repetition time (TR) for scanning was TR = 720 ms, with a spatial resolution of 2mm×2mm×2mm, and each brain volume had a shape of 91×109×91. All data have undergone minimal necessary preprocessing and have been registered to the Montreal Neurological Institute (MNI) 152 standard template space to eliminate individual structural differences [19], [20], [21]. Additionally, to remove blank areas in the data, each brain volume was cropped to 80×96×80.

The block-design task adopted in this study is the Motor task, designed by Bucker et al. [22]. The experiment utilized a block-design paradigm consisting of 6 different conditions: Left Hand, Right Hand, Left Foot, Right Foot, Tongue, and Cue. Each scan comprised 13 blocks, including 10 task blocks (two blocks per condition) and three 15-second baseline blocks. Each run lasted 284 volumes, and each participant was required to complete two scans using RL and LR phase encodings, respectively, with varying orders of task execution between the scans. The event-related design task employed the Gambling task, originally developed by Delgado et al. [23]. The task included 3 conditions: win, loss, and neutral, and was designed with a block-like structure, each run comprising 4 temporal blocks, including 2 main reward blocks and two main punishment blocks. Each temporal block consisted of 8 events, 6 of which were primary outcomes and 2 were other types of outcomes. The total duration of the scan was 253 volumes. Similar to Motor task, participants underwent 2 scans with different phase encodings, RL and LR, where the order of primary outcomes in the blocks varied. The correspondence between different conditions in the motor and gambling tasks regarding time and frame count is presented in Table I.

TABLE I
MOTOR TASK CONDITIONS AND LASTING TIME

| Task | States | Time |
|---|---|---|
| Block Design: Motor 284 TRs | left hand/ right hand/ left foot/ right foot/ tongue | 12s (17TRs) |
| | cue | 3s (4TRs) |
| | rest | 15s (21TRs) |
| Event Design: Gambling 253 TRs | win/ loss/ neutral | 3.5s (5TRs) |
| | rest | 15s (21TRs) |

### B. Model Architecture

Fig.1 illustrates the framework of the proposed tfMRI volume-wise decoder. The overall framework follows an encoder-decoder architecture. The encoder part is responsible for encoding multi-frame fMRI data into low-dimensional representations, aiming to capture abstract features and temporal dependencies in the data. the decoder part then decouples these features to obtain feature representations for each volume and ultimately infers the task state of each volume. The first layer of the encoder employs a 3D convolution layer with a kernel size of 1×1×1, which is here named the Time Embedding module, used to describe information in the temporal dimension and significantly reduce data dimensionality. The backbone network of the encoder utilizes a Residual Network, designed for efficient feature extraction. A channel attention module is added after each residual block to further enhance the extraction of temporal dimension information. The decoder adopts the strategy of spatiotemporal feature decoupling, separating high-dimensional features into temporal and spatial subspaces. Finally, the output classification task state of each volume is obtained through Softmax.

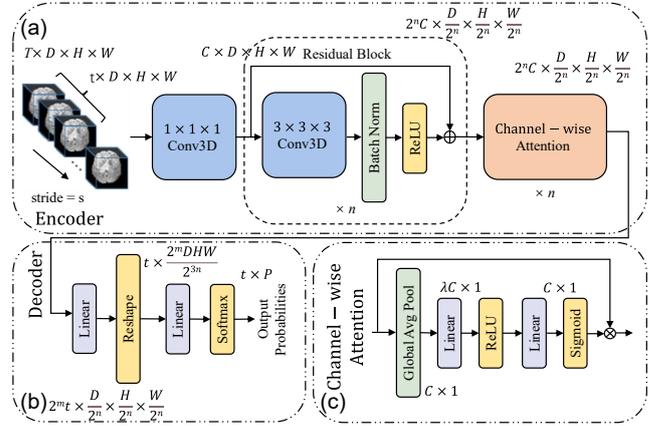

Fig.1. The classification model of the tfMRI based on ResNet. (a)The structure of the encoder based on ResNet. (b) The structure of the decoder. (c) The structure of channel attention.

#### 1) Time Embedding

The Time Embedding module integrates temporal dynamics into channel dimensions via a 1×1×1 3D convolution, mapping input data from $t \times D \times H \times W$ to $c \times D \times H \times W$. This preserves spatial structure while encoding temporal descriptors into channels, enhancing nonlinear representation and reducing computational demands [9]. During training, it dynamically adjusts weights to compress high-dimensional fMRI data (e.g., temporal frames t to channels c), enabling efficient spatiotemporal feature learning.

#### 2) Feature Extractor

Due to the characteristics of its filters, Convolutional Neural Networks (CNNs) primarily focuses on local data features. For commonly used convolutional kernels, such as 3×3×3, the network primarily focuses on local spatial information, which requires less computational space for each operation. In contrast, The Transformer architecture, due to its attention to global relationships among all tokens, requires larger computational space for each operation. Although parallel processing can speed up efficiency, the limitations in computational resources prevent Transformers from fully leveraging their advantages when training with smaller data batches. As a result, the Transformer efficiency is still inferior to the CNN architecture.

*3) Channel-wise Attention*

The Channel Attention mechanism we adopt is illustrated in Fig.1(c). The proposed module captures spatial-wise channel dependencies through multi-axis pooling, compressing fMRI volumes into $C \times 1$ feature vectors. These vectors are refined via bottleneck layers with nonlinear activation to generate attention scores. By broadcasting and multiplying the scores across the original 4D tensor, the model dynamically weights voxel-wise features, followed by normalization to prioritize task-relevant spatiotemporal patterns. This design enhances decoding precision while preserving computational efficiency through dimension-aware feature selection.

*4) 3D Swin Transformer*

We also tried replacing the backbone in the Fig.1 model with 3D Swin Transformer [24], while leaving the decoder part unchanged. The 3D Swin Transformer comprises four components: Patch Partition, Linear Embedding, 3D Swin Transformer Block, and Patch Merging. As shown in Fig.2(a), Stage 1 combines Linear Embedding with 3D Swin Transformer block. Subsequent stages (Stage 2-4) iteratively merge adjacent 2×2×2 token groups into 8C-dimensional features, linearly projected to 2C dimensions. This gradually reduces the resolution from D'/4 × H'/4 × W'/4 to D'/8 × H'/8 × W'/8, enabling multi-scale feature learning.

Local-window self-attention (Swin Transformer's core design) minimizes computation via window-based attention within partitioned volumes, with shifted windows enabling cross-window interaction for global context modeling. Fig.(b) shows the structure of the 3D Swin Transformer module. Combined with the moving window, two consecutive 3D Swin Transformer blocks can be calculated as:

$$\hat{z}^l = 3DW - MSA(LN(z^{l-1})) + z^{l-1}, \quad (1)$$
$$z^l = FFN(LN(\hat{z}^l)) + \hat{z}^l, \quad (2)$$
$$\hat{z}^{l+1} = 3DSW - MSA(LN(z^l)) + z^l, \quad (3)$$
$$z^{l+1} = FFN(LN(\hat{z}^{l+1})) + \hat{z}^{l+1}. \quad (4)$$

$z^{l-1}$ represents the input of Layer l 3D Swin Transformer block, $\hat{z}^l$ represents the output of 3D W-MSA in Layer l block, and $z^l$ represents the output of Layer l block. $\hat{z}^{l+1}$ represents the output of 3D SW-MSA in layer l+1 block, and $z^{l+1}$ represents the output of layer l+1 block.

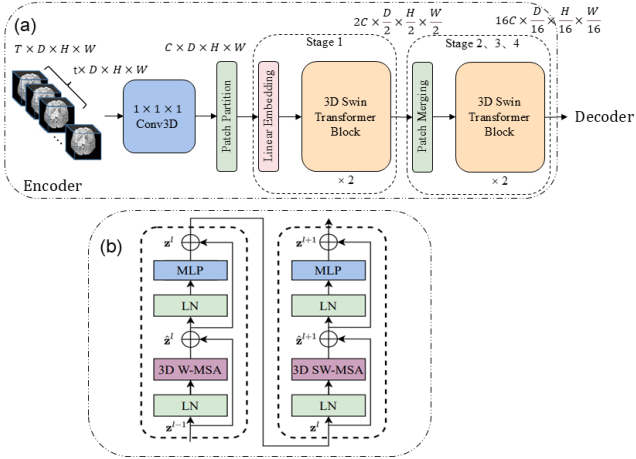

Fig.2. The classification model of the block-design tfMRI based on 3D Swin Transformer. (a) The structure of the encoder comprising the Time Embedding module and 3D Swin Transformer module; (b) The structure of the 3D Swin Transformer module.

### C. Parameters Settings and Experiment Setup

The model was implemented in PyTorch using tfMRI data from 1,080 subjects (motor and gambling tasks, RL/LR phase encoding). Data were split into training/validation/test sets (864/108/108 subjects) and spatially cropped to 80×96×80. To enhance robustness, a sliding window strategy ($t$=16 volumes) was adopted: during training, random temporal cropping generated augmented inputs, while inference used overlapping windows (stride $s$=1) with majority voting to resolve multi-prediction conflicts. To address HRF latency, labels were shifted by $l$=4 volumes. Training utilized the AdamW optimizer (batch size=16, weight decay=0.05) with Cross Entropy Loss. The learning rate warmed up linearly from 2e-5 to 2e-4 in 2 epochs, followed by cosine decay over 20 epochs. To mitigate memory constraints, each sequence was segmented into $n$ sub-samples per batch, accelerating training efficiency.

### D. Visualization

The Gradient-based visualization algorithm provides feedback on the gradient of each voxel in the input image at various time points for classification results, while maintaining image resolution. This study employs the gradient-based guided backpropagation (GBP) algorithm for visual decoding of fMRI data, focusing separately on motor and gambling tasks.

Using a multiple input and output approach during training and inference, the GBP algorithm generates gradient images with spatial dimensions and temporal length matching that of the input. To minimize overlap from simultaneous gradient outputs, only one volume is selected per round with a stride of 1. For an input length of 16, we retain only the 8th frame; initial and final frames representing rest periods are discarded as less meaningful. Thus, for motor tasks with an input length of 284 volumes, we obtain 269 volumes of gradient images; for gambling tasks with an input length of 253 volumes, we produce 238 volumes.

Averaging these gradient images yields a brain activation map across the entire time series for the group. GLM regression analysis then computes regression coefficients between each voxel and different states to measure correlation between voxels and states. By setting varying thresholds for each state, corresponding spatial maps can be obtained to accurately localize specific cognitive function areas in the brain. The time series from voxels output by GBP reflects temporal characteristics decoded by the model.

## III. RESULTS

### A. Block-Designed Task

Firstly, the performance of the decoding model based on ResNet is compared with the decoding model based on 3D Swin Transformer. Fig.3 shows the comparison of accuracy and F1 scores of the two models in frame-level classification of motor tasks. The ResNet-based model achieved an average accuracy of 94.0±8.2% and an F1 score of 94.0±7.2% (mean ± st.d.). This represents an improvement of 1.9 percentage points in accuracy and 2.2 percentage points in F1 score compared to the 3D Swin Transformer-based model.

Therefore, the subsequent analysis primarily focuses on the results of the model based on ResNet.

The experiment utilizes the Sparse Dictionary Learning (SDL) method proposed by Ge et al. [14] for comparison. This approach involves training and validation on the motor task subset of the HCP dataset and employs Pearson's Correlation Coefficient (PCC) as the evaluation metric to quantify the similarity between the classified result sequences and the ground truth sequences. To ensure consistency with the previous work, before calculating similarity, the standard task design sequences for the 6 task conditions and the predicted stimulus sequences were convolved with the canonical HRF to simulate the sequence of BOLD signal changes in the brain due to stimuli.

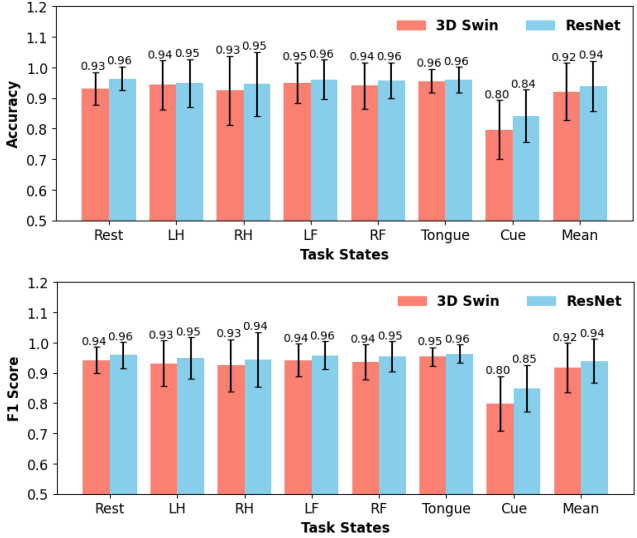

Fig.3. Comparison of accuracy and F1 score of two models in motor task frame classification.

Table II displays the comparative results of sequence similarity for 6 states decoded between our model and the compared model. It is evident from the table that the PCC values of the model introduced have improved by 0.20, 0.19, 0.15, 0.15, 0.18, and 0.31 respectively over the SDL method, with an average improvement of 0.19.

TABLE II
DECODING ACCURACY OF BLOCK-DESIGN tfMRI (MEAN±ST.D.)

| Task States | SDL | Ours |
|---|---|---|
| Left hand | 0.78±0.03 | **0.98±0.07** |
| Right hand | 0.78±0.03 | **0.97±0.07** |
| Left foot | 0.83±0.02 | **0.98±0.04** |
| Right foot | 0.83±0.01 | **0.98±0.05** |
| Tongue | 0.81±0.03 | **0.99±0.03** |
| Cue | 0.63±0.02 | **0.94±0.05** |
| Mean | 0.78±0.02 | **0.97±0.06** |

Fig.4 summarizes the motor task classification performance. The model achieved an average accuracy of 94.0% (±8.2%) and F1 score of 94.0% (±7.2%) across states, with highest accuracy for tongue movements (96.4%) and lowest for visual cues (84.2%). The confusion matrix (Fig. 4c) confirms strong diagonal dominance, indicating robust state discrimination. ROC analysis (Fig. 4d) further validates performance, showing near-perfect AUC for tongue (0.999) and high AUC for cues (0.980), consistent with accuracy trends.

The observed results stem from state-specific characteristics and task sequence design:

1) Visual cues in motor tasks exhibit brief durations and precursor roles. Temporal drift during detection (Fig.5) explains both their reduced accuracy and the elevated misclassification rates in the confusion matrix's final quadrant.

2) The "magnetic field equilibration" time at the start of fMRI data collection reduces data quality [13], and reduces the classification accuracy of the resting state. This is also the reason why a segment of data at the beginning of the run sequence is often discarded in numerous neuroimaging analyses.

3) Superior differentiation of distinct body parts versus left/right movements reflects bilateral motor cortex activation during unilateral tasks [25]. Tongue movements' unilateral nature enables peak accuracy. Fig.5 demonstrates precise temporal alignment between decoded and actual sequences, with state transitions typically deviating ≤2 TRs (1.5s), confirming the model's capacity to capture temporal patterns and HRF-modulated BOLD signals.

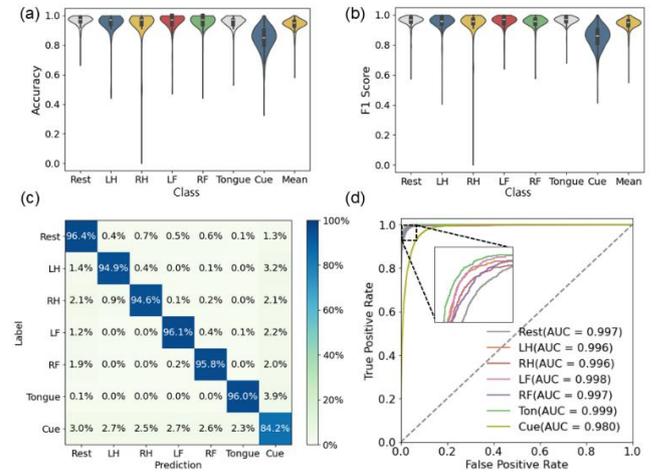

Fig.4. Frame classification results of motor tasks using the classification model based on ResNet. (a) The accuracy of each state (b) F1 scores of each state (c) Confusion matrix for each state accuracy; (d) Receiver Operating Characteristic (ROC) curve and Area Under the Curve (AUC) value of each state of frame classification.

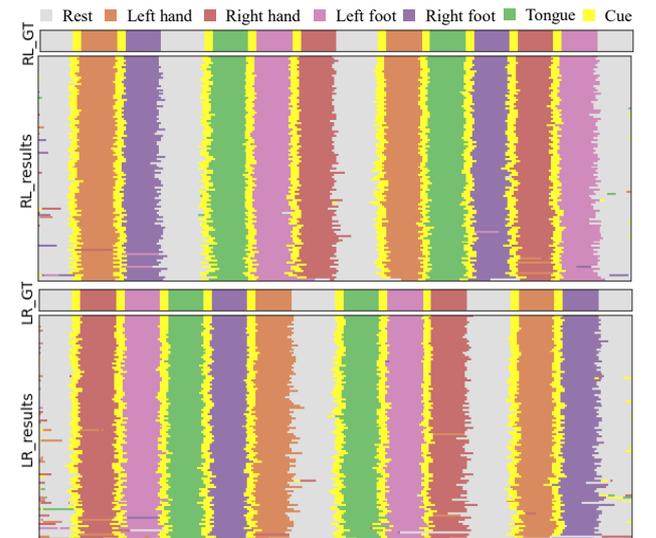

Fig.5. Individual results of motor task decoding sequences. From top to bottom is the truth value of RL phase sequence, the individual decoding result of RL phase sequence, the truth value of LR phase sequence, and the individual decoding result of LR phase sequence. The results of individual decoding shown in the Figure are arranged in descending order of accuracy.

The results presented above indicate that the model proposed in this study exhibits excellent performance in capturing the underlying spatiotemporal characteristics of fMRI signals. It enables fully automated, end-to-end frame classification for block-design task-based fMRI, achieving significant performance improvements compared to previous methods.

### B. Event-Related Task

Fig.6(a&b) display classification performance for gambling task states. The model achieved accuracies of 95.5±4.9% (rest), 71.8±18.0% (loss), 75.3±12.9% (win), and 36.4±25.9% (neutral) (mean 79.6±7.1%), with corresponding F1 scores of 94.7±4.5%, 70.9±12.5%, 73.4±9.7%, and 44.2±25.9% (mean 78.6±7.1%). All states significantly surpassed baseline performance.

Fig.6(c) presents the confusion matrix for the accuracy of classifying each task state by the model. Overall, each state achieved an accuracy rate higher than the baseline. Specifically, the findings include:

1) The rest state achieved better accuracy compared to the task states, indicating that the model first learned well the difference between tfMRI in gambling and resting-state fMRI.

2) Both loss and win states achieved accuracy rates exceeding 70%, with most false negatives occurring as misclassifications into the opposite state. These results demonstrate that the model is capable of distinguishing between reward-related and punishment-related brain functional activity patterns at the individual level, even in cross-subject settings.

3) Despite the fewer number of events and shorter duration of the neutral state in the stimulus sequence, the model still exhibited discriminative ability for the neutral state, with an accuracy rate above the baseline level. This result reflects the model's robustness in handling sparse and transient events. However, the relatively small sample size for the neutral state underscores potential avenues for improving classification performance, particularly with more balanced datasets.

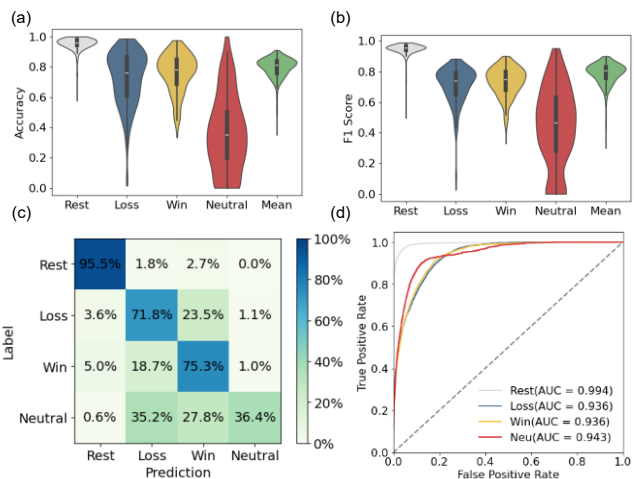

Fig.6. Classification results of gambling task frames. (a) The accuracy of each state of the frame classification subjects; (b) F1 scores of each state (c)Confusion matrix for each state accuracy (d) ROC curve and AUC value of each state.

Figure 6(d) shows the ROC curves for each state of the gambling task. Each state achieved a high AUC value, with the largest being the rest state at 0.994. However, the neutral state, which had the lowest classification accuracy, still achieved an AUC value as high as 0.943, even higher than that of win and loss. This is mainly due to the calculation method of the ROC curve; as mentioned earlier, the neutral state had fewer trial events, resulting in far more true negatives than false positives, thereby increasing the ROC value. However, this does not affect the overall evaluation of the model's performance, as the AUC value mainly reflects the comprehensive performance of the model under different classification thresholds.

Fig.7 demonstrates the individual results sequence for the classification of gambling task states. From the block level, the classification boundaries of each block are clear and align well with the actual stimulus sequence, indicating that the model effectively identifies transitions between blocks. However, classification results show a trend of deterioration over time. The data is divided into groups of two blocks based on the collection sequence: each participant's data includes the first half and second half of both RL and LR phase sequences. The classification accuracy rates for these segments are 84.2±8.1%, 77.5±9.3%, 80.5±7.0%, and 76.0±6.6%, reflecting decreases of 6.7% and 4.5% for each phase sequence respectively, with an average decrease in accuracy of 5.1% for LR compared to RL. Research by Abe et al. [26] suggests that during tasks like gambling, a sequential effect may occur when participants adapt or experience fatigue from repeated stimuli during fMRI scanning, leading to reduced response amplitudes, increased signal-to-noise ratios, decreased data quality, and potentially severe outcomes such as data distortion or experimental failure. In the gambling task, participants were exposed to multiple reward and punishment stimuli, progressively adapting to the stimuli over time. This adaptation resulted in a decline in the BOLD response and a corresponding downward trend in classification accuracy. This suggests the model's potential capability in analyzing sequential effects within scanning [27].

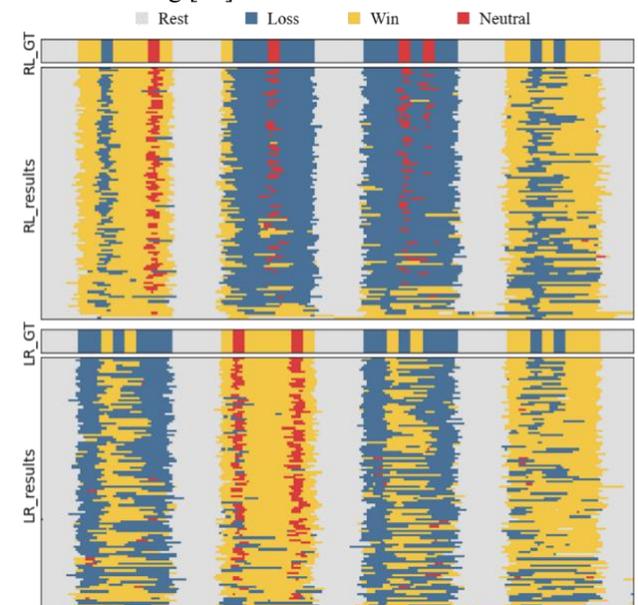

Fig.7. Sequence of individual results for gambling task state classification. From top to bottom is the truth value of RL phase sequence, the individual decoding result of RL phase sequence, the truth value of LR phase sequence, and the individual decoding result of LR phase sequence. The results of individual decoding shown in the Figure are arranged in descending order of accuracy.

At the event level, visualization results of neutral event classification show clear patterns, particularly in the first block of the RL phase sequence and second block of the LR phase sequence. This further confirms that while overall accuracy for neutral states is lower, the model still effectively detects neutral event stimuli among some participants.

The model effectively detects event switches between punishment stimuli in reward blocks and reward stimuli in punishment blocks, demonstrating its ability to differentiate between brain activity patterns related to reward and punishment. The detection accuracy at the end is influenced by subsequent events: if no identical or different types of events occur in the next two trials, as seen in the first two RL phase blocks, the model performs well. However, if the same type of event reappears within these trials, performance declines, as observed in other blocks. Due to the HRF characteristics of BOLD signals, after an event ends (within 3.6 seconds or 1 trial), BOLD responses remain elevated; thus, presenting a repeated stimulus can cause another spike in BOLD signal levels. This sustained high state complicates differentiation among intermediate events. This phenomenon contributes to inaccuracies noted in the confusion matrix for decoding various states during gambling tasks, where false negatives for rewards and punishments often misclassify into opposing states.

The results indicate that the model can capture the implicit temporal dependencies in short-time scale neural activity sequences, demonstrating the effectiveness of the model in decoding brain neural activities. Additionally, we discovered Sequential Effect that may occur in participants during the gambling task. This effect is an important reason for the downward trend in the model's classification level over time.

### C. Visualization

Motor task used the LR phase data from 108 participants in the test set as the source data for visualization analysis. Fig. 8 presents group-level spatial visualization results of the motor task, with a higher threshold applied due to strong mapping activation. In Fig. 8 (a-f), all six states show significant brain region mappings. Visual cue activation is primarily located in the occipital visual area, while activations for the left hand, right hand, left foot, right foot, and tongue are concentrated in the motor cortex. Foot activation occurs near the midline of the contralateral parietal lobe's motor cortex; hand activation is ventral to foot activation on the contralateral side; and tongue activation is ventral to bilateral hand activations. Similar findings were observed in the cerebellum, aligning with known topological characteristics of motor functions [28]. This indicates that the model effectively focuses on biologically relevant regions and learns their features for high-precision volume-wise classification of task states.

Negative activations showed cross-task consistency (e.g., Fig.8(g) occipital deactivation during left-hand movements), arising from BOLD signal decline at visual cue termination, which is a temporal contrast feature leveraged for state classification. This demonstrates the deep learning model's data-driven decoding prioritizes task-relevant signal dynamics over strict neuroanatomical modularity, capturing both canonical activations and transitional hemodynamic patterns.

Selecting the voxel with the highest regression coefficient from each state's brain region shown in Fig.9, flatting it over the time dimension, and plotting it together with the stimulus sequence and the sequence obtained by convolving the stimulus sequence with the standard HRF results in the curve chart shown in Fig.9. From the figure, the time series obtained from the model visualization and the standard convolution result sequence are almost completely consistent in trend. At the beginning of the task state stimulus, both begin to rise, and their rising trajectories almost coincide; as the stimulus continues, both maintain an upward trend; when the task stimulus ends, both reach their peak and then begin to decline. This high degree of consistency indicates that the model has extremely high temporal sensitivity to block-designed tfMRI, capable of precisely capturing changes in brain activity at state transitions, and can also effectively capture and depict the HRF characteristics of the BOLD signal.

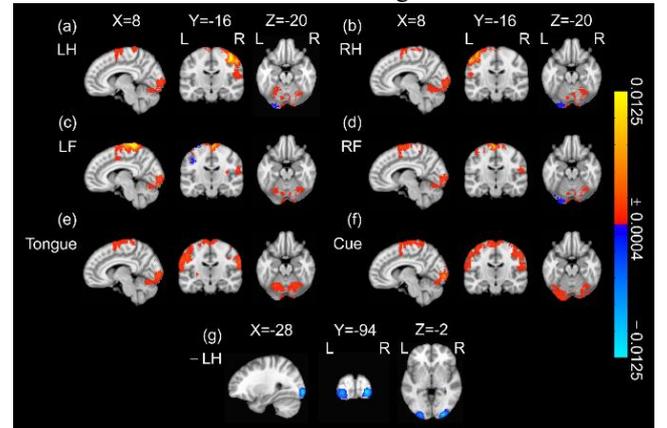

Fig.8. Spatial visualization results of motor task decoding. (a-f) Activation maps for left/right hands, left/right feet, tongue, and visual cues; (g) left hand negative activation. The right ColorBar value is the regression coefficient (FDR-corrected p＜10-15).

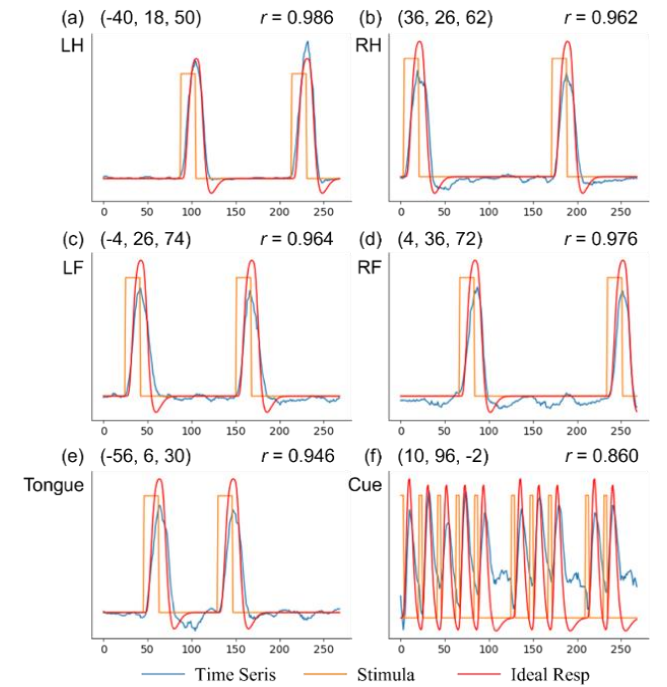

Fig.9. Time visualization results of motor task decoding. (a-f) Voxel time series for left/right hand, left/right foot, tongue, and visual cue regions. Blue: decoded signal; yellow: stimulus timing; red: ideal BOLD response. MNI coordinates with PCC values (all p＜10-10) are labeled.

From the above results, it is known that gambling task may cause a decline in recognition accuracy over time due to Sequential Effect. Therefore, this section only uses the data from the first two blocks of the RL phase of the 108 participants in the test set as the source data for visualization analysis. Similarly, the following sections will present the visualization analysis of the model's decoding the gambling task at both the spatial and temporal levels.

Fig.10 shows the spatial visualization results of gambling task. It can be seen that the striatum and insula related to rewards appear in the activated brain regions, and areas related to vision also show strong activation, yielding conclusions similar to those of the GLM analysis results from the HCP [28]. Additionally, Fig.8(a-c) display that the activation areas for loss, win, and neutral tend to be the same. The reason is that in the main event blocks, there is a rapid alternation of task states over a short period, and the BOLD response is of a long-time scale, causing intermediate state values to always maintain a high level, thereby leading to confusion between states.

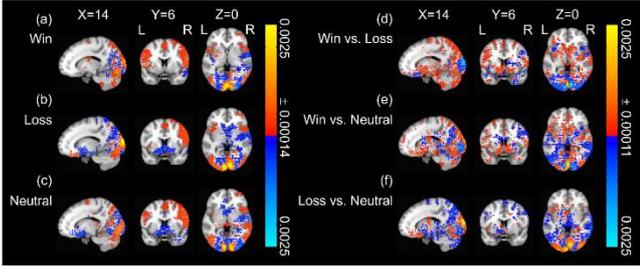

Fig.10. Spatial visualization results of gambling task decoding. (a-c) Activation maps for win, loss, and neutral states; (d) win vs loss; (e) win vs neutral; (f) loss vs neutral. The right ColorBar value is the regression coefficient (FDR-corrected $p<10^{-4}$).

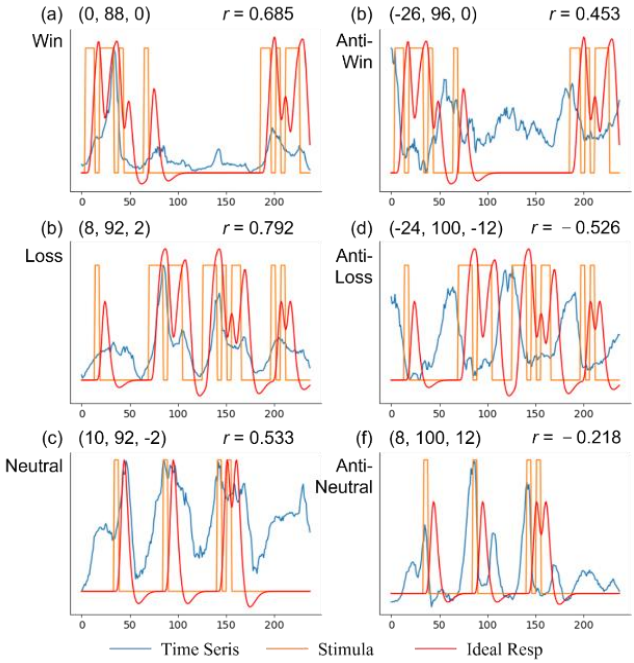

Fig.11. Time visualization results of gambling task decoding. (a-f) Voxel time series for win (±), loss (±), and neutral (±) activation. Blue: decoded signal; yellow: stimulus timing; red: ideal BOLD response. MNI coordinates with PCC values (all $p<10^{-10}$) are labeled.

Based on this, the experiment explored the pairwise comparison results of task states. As shown in Fig.10(d-f), the activation is mainly concentrated in the visual area, with the difference lying in the distribution of regions and the numerical distribution for different contrasts.

For gambling task, a temporal visualization analysis was conducted on 3 cognitive states: win, loss, and neutral. Given that these 3 states correspond to brain regions with a wide range of negative activation areas, this section also selected the voxels with the lowest regression coefficients among these 3 states for analysis. The results are shown in Fig.11, where the PCCs for the positive activation voxels of reward, punishment, and neutrality are 0.792, 0.685, and 0.533, respectively, and the PCCs for the negative activation voxels are -0.453, -0.526, and -0.218, respectively, all with p-values less than $10^{-3}$. It can be seen that both punishment and reward responses have higher correlation coefficients with the ideal response, especially their positive activation time series, which still show convergence with the standard convolution result sequence at certain points in time. This indicates that the model has initially learned the activity patterns of rewards and punishments and can distinguish between them, suggesting that it is potential for volume-wise decoding of event-related tfMRI via deep neural network for development and improvement.

## IV. DISCUSSION AND FUTURE WORK

We proposed a volume-wise classifier for tfMRI based on DNN, and achieved remarkable results in motor task and gambling task. On the one hand, this work provided a direct basis for the explainability and biological significance of the internal learning patterns of the model; on the other hand, it offered a finer spatial division of brain regions mapping cognitive functions in different states spatially, and revealed the dynamic changing characteristics of brain activities temporally, providing a new perspective for the dynamic research of cognitive functions.

Although we have conducted a relatively comprehensive exploration of tfMRI volume-wise decoding based on deep learning, there are still some limitations and open issues. Firstly, while we achieved promising results, there is still room for improvement in the model's decoding capabilities, especially in the detection of continuous rapid switching events in event-related design tasks. To further optimize classification performance, improvements to the network structure could be considered, such as enhancing the decoder module to extract temporal dependence features, or designing more refined strategies for decoding task-state data in event-related designs. We also tried using 3D Swin Transformer as the backbone network for the model, but it was still slightly less effective than ResNet. We believe that the reasons for the performance difference may be as follows:

1) While Swin Transformer reduces the computational demands of ViT [29], its memory intensity persists in high-dimensional fMRI processing. Computational constraints necessitate dimensionality reduction via time/linear embedding modules, inducing information loss that limits both model potential exploitation [30] and decoding accuracy through compromised spatial-temporal resolution.

2) Transformers' global modeling capacity [31] proves suboptimal for fMRI's regionally clustered activations, particularly given typically limited datasets from high acquisition costs [32]. The localized nature of cognitive neural patterns further diminishes Transformer advantages compared to CNNs' efficient local feature extraction in small-data regimes.

Secondly, designing and developing a system for tfMRI volume-wise decoding based on deep learning would be a potentially promising direction. This approach could potentially uncover more intricate neural patterns within the brain, enabling a more accurate understanding of the underlying cognitive processes. By leveraging the powerful feature - extraction capabilities of deep neural networks, the system might be able to distinguish between different mental states with a higher degree of precision than traditional methods. Lastly, the data used in this study were mainly from large-scale tfMRI data from the single dataset, with relatively single data patterns. Future research could consider methods such as large-scale pre-training and transfer learning to incorporate resting-state fMRI data and small cross-site datasets into analysis, aiming to build an easy, interpretable, and more generalizable deep learning model, or even construct a multimodal neural imaging large model. This will greatly promote the promotion of deep learning models in practical applications and facilitate the development of brain science.

## V. CONCLUSION

In this paper, we proposed a tfMRI volume-wise classifier based on DNN. Cross-subject validation was performed on both a block-designed motor task and an event-related gambling task. Results indicated that the model significantly enhanced volume classification performance for both tasks, successfully capturing potential spatiotemporal relationships in tfMRI data, identifying neural activities across different time scales, and improving temporal resolution from block-level to volume-level. Finally, cognitive patterns from motor and gambling task data were visualized using visualization algorithms.

## REFERENCES

[1] M. Hanke, Y. O. Halchenko, P. B. Sederberg, S. J. Hanson, J. V. Haxby, and S. Pollmann, 'PyMVPA: a Python Toolbox for Multivariate Pattern Analysis of fMRI Data', *Neuroinformatics*, vol. 7, no. 1, pp. 37–53, Mar. 2009, doi: 10.1007/s12021-008-9041-y.

[2] M. N. Hebart, K. Görgen, and J.-D. Haynes, 'The Decoding Toolbox (TDT): a versatile software package for multivariate analyses of functional imaging data', *Front. Neuroinformatics*, vol. 8, p. 88, 2014, doi: 10.3389/fninf.2014.00088.

[3] N. N. Oosterhof, A. C. Connolly, and J. V. Haxby, 'CoSMoMVPA: Multi-Modal Multivariate Pattern Analysis of Neuroimaging Data in Matlab/GNU Octave', *Front. Neuroinformatics*, vol. 10, p. 27, 2016, doi: 10.3389/fninf.2016.00027.

[4] M. S. Treder, 'MVPA-Light: A Classification and Regression Toolbox for Multi-Dimensional Data', *Front. Neurosci.*, vol. 14, p. 289, 2020, doi: 10.3389/fnins.2020.00289.

[5] G. H. Glover, 'Overview of functional magnetic resonance imaging', *Neurosurg. Clin. N. Am.*, vol. 22, no. 2, pp. 133–139, vii, Apr. 2011, doi: 10.1016/j.nec.2010.11.001.

[6] L. Song, Y. Ren, S. Xu, Y. Hou, and X. He, 'A hybrid spatiotemporal deep belief network and sparse representation-based framework reveals multilevel core functional components in decoding multitask fMRI signals', *Netw. Neurosci.*, vol. 7, no. 4, pp. 1513–1532, Dec. 2023, doi: 10.1162/netn_a_00334.

[7] N. Xi, S. Zhao, H. Wang, C. Liu, B. Qin, and T. Liu, 'UniCoRN: Unified Cognitive Signal ReconstructioN bridging cognitive signals and human language', Jul. 06, 2023, *arXiv*: arXiv:2307.05355. doi: 10.48550/arXiv.2307.05355.

[8] S. Gupta, M. Lim, and J. C. Rajapakse, 'Decoding task specific and task general functional architectures of the brain', *Hum. Brain Mapp.*, vol. 43, no. 9, pp. 2801–2816, 2022, doi: 10.1002/hbm.25817.

[9] X. Wang et al., 'Decoding and mapping task states of the human brain via deep learning', *Hum. Brain Mapp.*, vol. 41, no. 6, pp. 1505–1519, Apr. 2020, doi: 10.1002/hbm.24891.

[10] S. Zhao et al., 'Task sub-type states decoding via group deep bidirectional recurrent neural network', *Med. Image Anal.*, vol. 94, p. 103136, May 2024, doi: 10.1016/j.media.2024.103136.

[11] F.-C. Su, T.-C. Chu, Y.-Y. Wai, Y.-L. Wan, and H.-L. Liu, 'Temporal resolving power of perfusion- and BOLD-based event-related functional MRI', *Med. Phys.*, vol. 31, no. 1, pp. 154–160, 2004, doi: 10.1118/1.1634480.

[12] S.-G. Kim, W. Richter, and K. Uğurbil, 'Limitations of temporal resolution in functional MRI', *Magn. Reson. Med.*, vol. 37, no. 4, pp. 631–636, 1997, doi: 10.1002/mrm.1910370427.

[13] R. A. Poldrack, J. A. Mumford, and T. E. Nichols, *Handbook of Functional MRI Data Analysis*. Cambridge: Cambridge University Press, 2011. doi: 10.1017/CBO9780511895029.

[14] B. Ge, H. Wang, P. Wang, Y. Tian, X. Zhang, and T. Liu, 'Discovering and characterizing dynamic functional brain networks in task FMRI', *Brain Imaging Behav.*, vol. 14, no. 5, pp. 1660–1673, Oct. 2020, doi: 10.1007/s11682-019-00096-6.

[15] P. S. Scotti et al., 'Reconstructing the Mind's Eye: fMRI-to-Image with Contrastive Learning and Diffusion Priors', Oct. 07, 2023, *arXiv*: arXiv:2305.18274. Accessed: Jan. 13, 2024. [Online]. Available: http://arxiv.org/abs/2305.18274

[16] A. Veloz et al., 'Fuzzy General Linear Modeling for Functional Magnetic Resonance Imaging Analysis', *IEEE Trans. Fuzzy Syst.*, vol. 28, no. 1, pp. 100–111, Jan. 2020, doi: 10.1109/TFUZZ.2019.2936807.

[17] A. Cortese et al., 'The DecNef collection, fMRI data from closed-loop decoded neurofeedback experiments', *Sci. Data*, vol. 8, no. 1, p. 65, Feb. 2021, doi: 10.1038/s41597-021-00845-7.

[18] H. Wang et al., 'Recognizing Brain States Using Deep Sparse Recurrent Neural Network', *IEEE Trans. Med. Imaging*, vol. 38, no. 4, pp. 1058–1068, Apr. 2019, doi: 10.1109/TMI.2018.2877576.

[19] M. F. Glasser et al., 'The minimal preprocessing pipelines for the Human Connectome Project', *NeuroImage*, vol. 80, pp. 105–124, Oct. 2013, doi: 10.1016/j.neuroimage.2013.04.127.

[20] D. C. Van Essen et al., 'The WU-Minn Human Connectome Project: an overview', *NeuroImage*, vol. 80, pp. 62–79, Oct. 2013, doi: 10.1016/j.neuroimage.2013.05.041.

[21] D. C. Van Essen et al., 'The Human Connectome Project: a data acquisition perspective', *NeuroImage*, vol. 62, no. 4, pp. 2222–2231, Oct. 2012, doi: 10.1016/j.neuroimage.2012.02.018.

[22] R. L. Buckner, F. M. Krienen, A. Castellanos, J. C. Diaz, and B. T. T. Yeo, 'The organization of the human cerebellum estimated by intrinsic functional connectivity', *J. Neurophysiol.*, vol. 106, no. 5, pp. 2322–2345, Nov. 2011, doi: 10.1152/jn.00339.2011.

[23] M. R. Delgado, L. E. Nystrom, C. Fissell, D. C. Noll, and J. A. Fiez, 'Tracking the hemodynamic responses to reward and punishment in the striatum', *J. Neurophysiol.*, vol. 84, no. 6, pp. 3072–3077, Dec. 2000, doi: 10.1152/jn.2000.84.6.3072.

[24] Z. Liu et al., 'Video Swin Transformer', presented at the Proceedings of the IEEE/CVF Conference on Computer Vision and Pattern Recognition, 2022, pp. 3202–3211. Accessed: Dec. 11, 2024. [Online]. Available: https://openaccess.thecvf.com/content/CVPR2022/html/Liu_Video_Swin_Transformer_CVPR_2022_paper.html

[25] D. A. Barany et al., 'Primary motor cortical activity during unimanual movements with increasing demand on precision', *J. Neurophysiol.*, vol. 124, no. 3, pp. 728–739, Sep. 2020, doi: 10.1152/jn.00546.2019.

[26] N. Abe, R. Nakai, K. Yanagisawa, T. Murai, and S. Yoshikawa, 'Effects of sequential winning vs. losing on subsequent gambling behavior: analysis of empirical data from casino baccarat players', *Int. Gambl. Stud.*, vol. 21, no. 1, pp. 103–118, Jan. 2021, doi: 10.1080/14459795.2020.1817969.

[27] A. J. Yu and J. D. Cohen, 'Sequential effects: Superstition or rational behavior?', *Adv. Neural Inf. Process. Syst.*, vol. 21, pp. 1873–1880, 2008.

[28] D. M. Barch et al., 'Function in the human connectome: task-fMRI and individual differences in behavior', *NeuroImage*, vol. 80, pp. 169–189, Oct. 2013, doi: 10.1016/j.neuroimage.2013.05.033.

[29] Z. Liu et al., 'Swin Transformer: Hierarchical Vision Transformer using Shifted Windows', Aug. 17, 2021, *arXiv*: arXiv:2103.14030. doi: 10.48550/arXiv.2103.14030.

[30] A. Dosovitskiy et al., 'An Image is Worth 16x16 Words: Transformers for Image Recognition at Scale', Jun. 03, 2021, *arXiv*: arXiv:2010.11929. doi: 10.48550/arXiv.2010.11929.

[31] A. Vaswani et al., 'Attention Is All You Need', Aug. 02, 2023, *arXiv*: arXiv:1706.03762. doi: 10.48550/arXiv.1706.03762.

[32] K. Dadi et al., 'Benchmarking functional connectome-based predictive models for resting-state fMRI', *NeuroImage*, vol. 192, pp. 115–134, May 2019, doi: 10.1016/j.neuroimage.2019.02.062.